\documentclass[letterpaper, 10 pt, journal, twoside]{ieeetran}  

\usepackage{amssymb,graphicx,amsmath,color,algorithm,algorithmic,url}
\usepackage[font=small,labelfont=bf]{caption}
\usepackage[caption=false,font=footnotesize]{subfig}
\usepackage[english]{babel}
\usepackage{blindtext}
\usepackage{afterpage}
\usepackage[noadjust]{cite} 
\usepackage{hhline,caption}
\usepackage{hyperref}
\usepackage{placeins}
\usepackage{booktabs}
\usepackage{multirow}
\usepackage{setspace} 
\usepackage{xcolor}
\usepackage{soul}
\sethlcolor{green}

\IEEEoverridecommandlockouts                              
\title{
\color{black}{TacFinRay: Soft Tactile Fin-Ray Finger with Indirect Tactile Sensing for Robust Grasping}
}

\author{Saekwang Nam$^{1}\textsuperscript{\textdagger}$, Bowen Deng$^{2}\textsuperscript{\textdagger}$, Loong Yi Lee$^{2}$, Jonathan M. Rossiter$^{2}$, Nathan F. Lepora$^{2*}$
    \thanks{The work is supported by the Royal Society's ISPF grant ``MultiTip: Advancing Robot Dexterity with Multimodal Vision-Based Tactile Sensing'' (ICA/R1/241213), a UKRI award ``Made Smarter Innovation - Research Centre for Smart, Collaborative Industrial Robotics'' (EP/V062158/1), a National Research Foundation of Korea (NRF) grant funded by the Korea government (MSIT) (No. RS-2024-00436182) and by an IITP grant funded by MSIT (IITP-2025-RS-2024-00437756, RS-2025-02263277). $^*$Corresponding author: Nathan F. Lepora ({\rm\footnotesize n.lepora@bristol.ac.uk}), \textsuperscript{\textdagger} Denotes equal contribution.}
    \thanks{$^{1}$Saekwang Nam is with the Graduate School of Data Science at Kyungpook National University, 41566 Daegu, Republic of Korea. Email: {\rm\footnotesize s.nam@knu.ac.kr}.}
    \thanks{$^{2}$Bowen Deng, Loong Yi Lee, Jonathan M. Rossiter and Nathan F. Lepora are with the University of Bristol, BS8 1QU Bristol, U.K., and the Bristol Robotics Laboratory, BS16 1QY Bristol, U.K.}

}

\begin{document}
\maketitle

\begin{abstract}
We present a tactile-sensorized Fin-Ray finger that enables simultaneous detection of contact location and indentation depth through an indirect sensing approach. A hinge mechanism is integrated between the soft Fin-Ray structure and a rigid sensing module, allowing deformation and translation information to be transferred to a bottom crossbeam upon which are an array of marker-tipped pins based on the biomimetic structure of the TacTip vision-based tactile sensor. Deformation patterns captured by an internal camera are processed using a convolutional neural network to infer contact conditions without directly sensing the finger surface. The finger design was optimized by varying pin configurations and hinge orientations, achieving 0.1\,mm depth and 2\,mm location-sensing accuracies. The perception demonstrated robust generalization to various indenter shapes and sizes, which was applied to a pick-and-place task under uncertain picking positions, where the tactile feedback significantly improved placement accuracy. Overall, this work provides a lightweight, flexible, and scalable tactile sensing solution suitable for soft robotic structures where the sensing needs situating away from the contact interface.

\end{abstract}


\section{INTRODUCTION}

Tactile sensing is essential for achieving dexterous manipulation in robotic hands~\cite{Bhirangi23RAL_tactile_sensing, zhang2025soft}. For example, to perform delicate tasks like gently grasping and placing eggs or glass plates, humanoid robots such as Figure's F.02 and Tesla's Optimus will need fingertip-mounted tactile sensors to become truly capable~\cite{Lepora24SciRobotics_RobotHand}. To enhance robotic dexterity, researchers have developed vision-based tactile sensors (VBTSs) that take advantage of recent advancements in computer vision~\cite{Faris23_SoRo, Lepora21Sensors_TacTipReview, Sferrazza19Sensors_OTS, Yuan15ICRA_SlipShear}. Unlike traditional tactile sensors that rely on direct physical interaction with the object~\cite{Fishel12BioRob_Biotac}, VBTSs utilize an internal mini-camera located to internally view the point of contact. This approach of decoupling the sensing module from the contact point enables significant sensor deformation while preserving high spatial resolution, making VBTS a powerful tool for tactile sensing in robotics~\cite{Lloyd24TacServoing_IJRR, Lambeta20DIGIT_RAL}.

A prominent example is the TacTip sensor, developed at the University of Bristol and Bristol Robotics Laboratory, which employs an internal camera to track the shear movement of marker-tipped pins embedded beneath a compliant sensor skin~\cite{Susini25_RAL, Zhang24_Tacpalm, Lepora21Sensors_TacTipReview, Li24_BioTacTip_RAL}. Similarly, MIT's GelSight sensor employs photometric stereo techniques to precisely reconstruct the surface geometry of an object at the point of contact~\cite{Yuan15ICRA_SlipShear,Wang24_GelSightFinray}. These sensors are increasingly acknowledged as fundamental to achieving dexterous robotic manipulation. By combining high-resolution visual feedback with physical interaction, VBTSs offers significant improvements over conventional tactile sensors, positioning them as a leading technology in robotic manipulation.

    \begin{figure}[t]
      \centering
      \vspace{1ex}
      \includegraphics[width = 0.80\columnwidth]{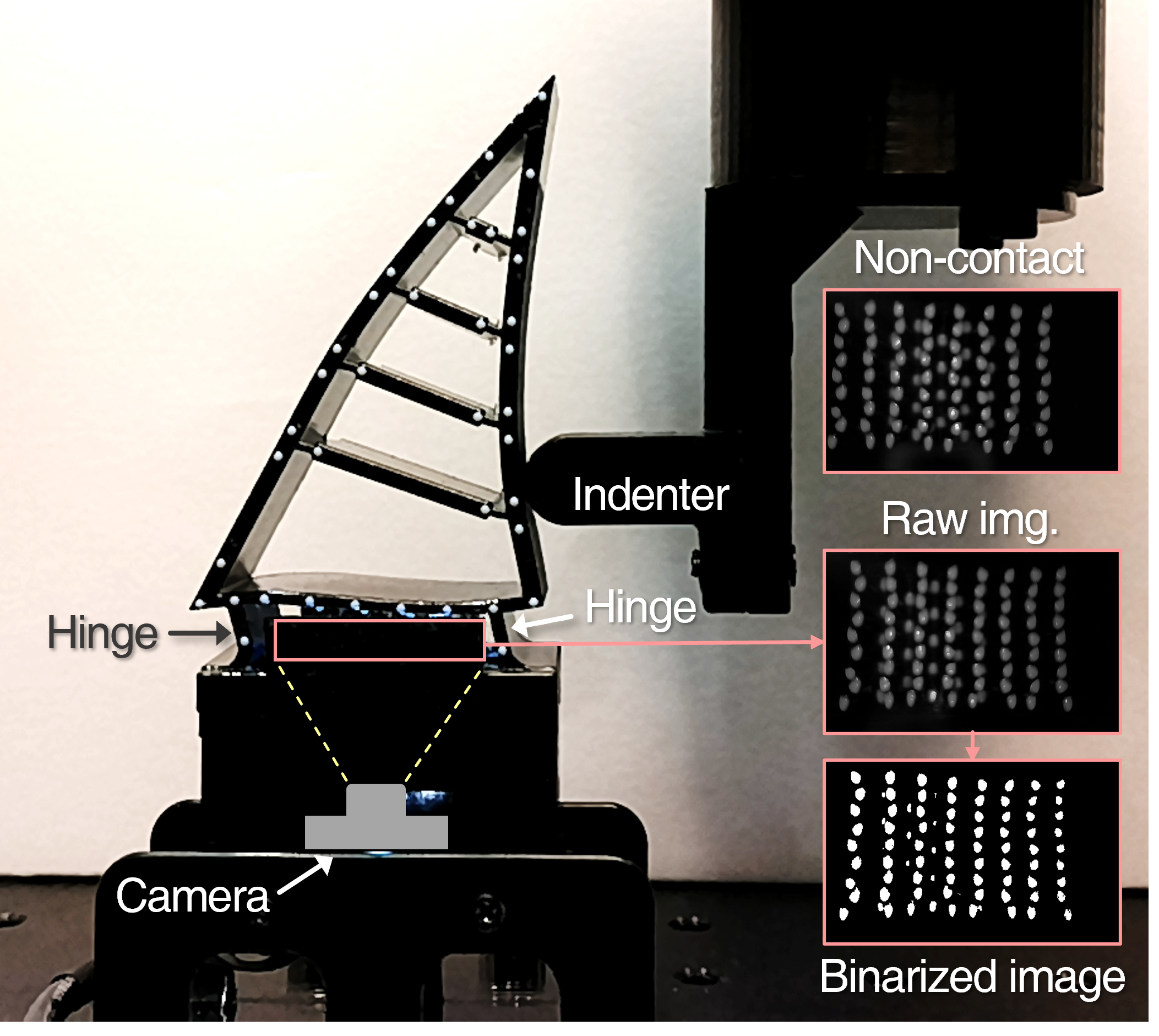}
      \caption{Proposed tactile Fin-Ray finger and experimental setup for validation. When physical contact occurs on the front surface, both the finger structure and the hinges undergo deformation. This deformation is represented in the marker-pin array on the bottom crossbeam. A camera, fixed at the base of the finger, captures the marker movement (inset images), which is processed as a binarized array for input to a contact location and depth prediction model.}
      \label{fig:Intro}
   \end{figure}

Despite the above-mentioned advantages of VBTSs, several limitations remain. First, reducing the overall thickness of the sensor is difficult due to the presence of rigid components such as the camera module and circuit board. As these components cannot be separated, a fixed volume behind the sensing skin must be allocated to accommodate them. To obtain a clear image of contacts, the distance between the sensing skin and the camera must be carefully adjusted, which increases the thickness of the sensor module. Second, achieving large-area sensing remains a challenge. The sensing area of a single VBTS is restricted by the field of view of its internal camera. While a study has attempted to expand the sensing area using reflective mirrors~\cite{Liu23RoboSoft_GelsighBabayFinray}, this approach introduces additional space requirements and complicates the design. Lastly, placing the camera far from the sensing skin is difficult, as the camera must capture the deformation of the skin upon contact. If the camera could be positioned away from the sensing area while still allowing tactile sensing, the thickness of the sensor could be significantly reduced.

In this paper, we present the development of a sensorised Fin-Ray finger that overcomes these limitations of VBTSs~(Fig.~\ref{fig:Intro}). Our proposed gripper exploits the inherent deformation of the Fin-Ray structure upon physical contact, facilitating indirect sensing by decoupling the contact interface from the deformation sensing unit. Furthermore, by introducing a hinge mechanism between the soft finger and the rigid sensing module, our approach enables the measurement of both contact location and indentation depth.

The main contributions of this paper are as follows:

\noindent 1) \textbf{Introduction of Hinge Mechanism}:
A hinge between the Fin-Ray structure and the rigid sensor module induces translation upon contact, enabling simultaneous measurement of contact location and depth. This provides a mathematical basis for 2D sensing by tracking the bottom crossbeam's movement (Sec.~\ref{sec:HingeMechanism}).

\noindent 2) \textbf{Indirect Crossbeam Sensing}: 
Our indirect method measures the deformation of the bottom crossbeam, which is transferred from the tactile surface. This approach decouples the sensing and contact areas, allowing greater design flexibility for the contact interface (Fig.~\ref{fig:Intro}).
  
\noindent 3) \textbf{Demonstration of Tactile Sensing Performance}:
Using a Convolutional Neural Network, we demonstrate precise detection of contact location and depth for various contact shapes (Sec.~\ref{sec:PosErr}). By comparing performance with and without sensing, we highlight the importance of tactile feedback for precise object manipulation.

\section{Related Work}

\subsection{Flexible Grippers with Tactile Sensing}
Recent studies embed tactile sensors into flexible grippers to enhance dexterity. A common method attaches sensors to contact areas, connecting rigid sensor components with soft joints to create a sensitive, flexible design~\cite{Lu22RAL_FourFinger}. Others employ vision-based tactile sensing to improve low spatial resolution~\cite{Lu24TMech_DexiTac}. However, because these sensors are built on rigid substrates, they introduce unavoidable rigidity in interaction areas, which can hinder stable gripping.
 
To address this limitation, fully soft grippers have been proposed~\cite{Wang24_GelSightFinray}. For instance, Guo et al.\,used a flexible Fin-Ray structure with embedded ArUco markers, which a camera tracks to measure contact location and deformation~\cite{Guo24AIS_FinRayARUCO}. This arrangement allows a camera to track the six-dimensional poses of the markers, enabling measurement of the contact location on the gripper and its deformation state~\cite{Guo24TRO_StateEstimate, Wang24_GelSightFinray, Liu23RoboSoft_GelsighBabayFinray}. A key drawback is that detecting markers at varying depths requires removing the finger's inner crossbeams. This compromises the structure's torsional stiffness and can diminish gripping performance.

\subsection{Indirect Sensing Methods}
Fin-Ray fingers' adaptive deformation makes them a focus of research for indirectly estimating contact or force. Shan and Birglen proposed a mathematical model to estimate gripping force from gripper deformation, but precise model parameter selection remains a significant challenge~\cite{Shan20IJRR_FinRayEffect}. To overcome this, another study used a neural network trained on deformation data from an external camera and forces from finite element model simulations~\cite{Xu21TRO_FinRayForce}. Other research has replaced external cameras with deflection sensors mounted on the finger's back to estimate deformation~\cite{Chen23TRO_DeflectionSensing}. Additionally, Yao et al.\,optimized the Fin-Ray design using genetic algorithms to enhance adaptability and grasping performance~\cite{Yao24MMT_FinRayEffect2}.

\subsection{Grippers Integrated with GelSight Sensors}

Numerous studies have proposed using GelSight sensors to enhance the tactile sensitivity of Fin-Ray grippers. The GelSight Fin Ray, for instance, incorporates a GelSight sensor at the Fin-Ray gripper's finger pad, using its high-resolution feedback to reconstruct 3D tactile deformations~\cite{Wang24_GelSightFinray}. However, the rigid GelSight module limits the structure's full conformal capability. Similarly, the GelSight Baby Fin Ray’s mirror slightly limits flexibility~\cite{Liu23RoboSoft_GelsighBabayFinray}, whereas the mirrorless TacFinRay ensures superior compliance. Table~\ref{tab:comparison} highlights TacFinRay's advantages over the GelSight Baby Fin Ray and a Fin Ray with direct sensing~\cite{10255383}. More recently, PneuGelSight was introduced, which consists of pneumatic actuators equipped with GelSight sensors, but its focus was on estimating gripper deformation rather than enhancing grasping performance~\cite{Zhang25_PneuGelSight}.

  \begin{table}[t]
      \centering
      \caption{Comparison of three indirect tactile Fin-Ray grippers}
      \label{tab:comparison}
      \includegraphics[width = 1.0\columnwidth]{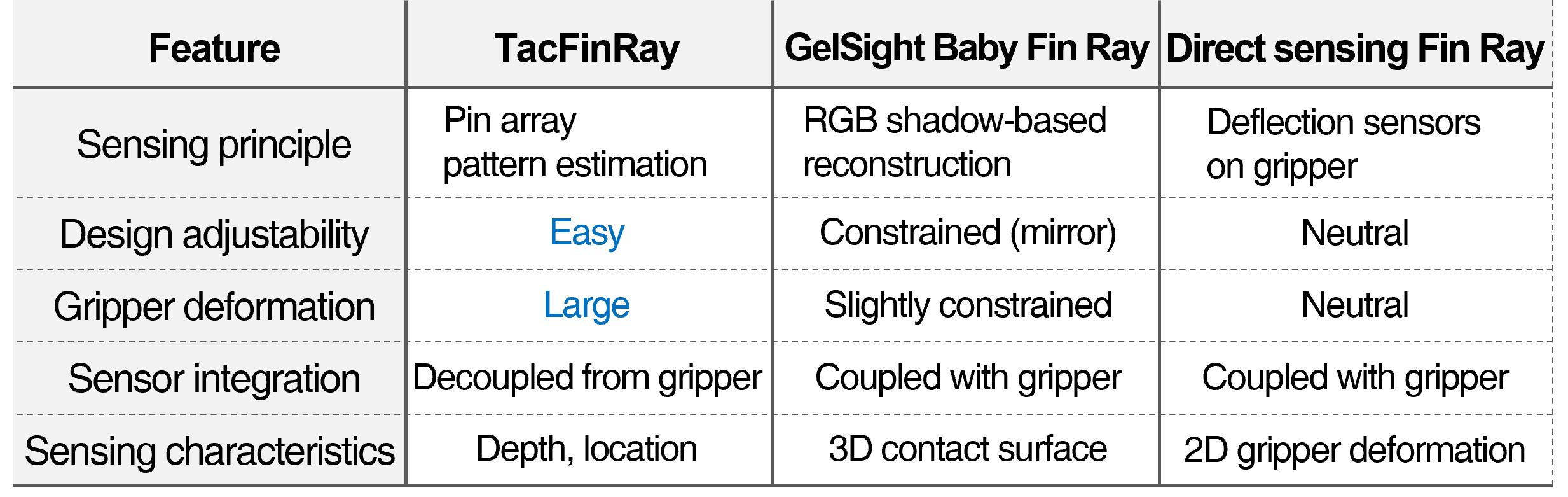}
  \end{table}




\section{Methodology}
\label{sec:methods}

\subsection{Fin-Ray Finger Design}
\subsubsection{Flexible 3D Printing}
\begin{figure}[t]
      \centering
      \vspace{1ex}
      \includegraphics[width = 0.9\columnwidth]{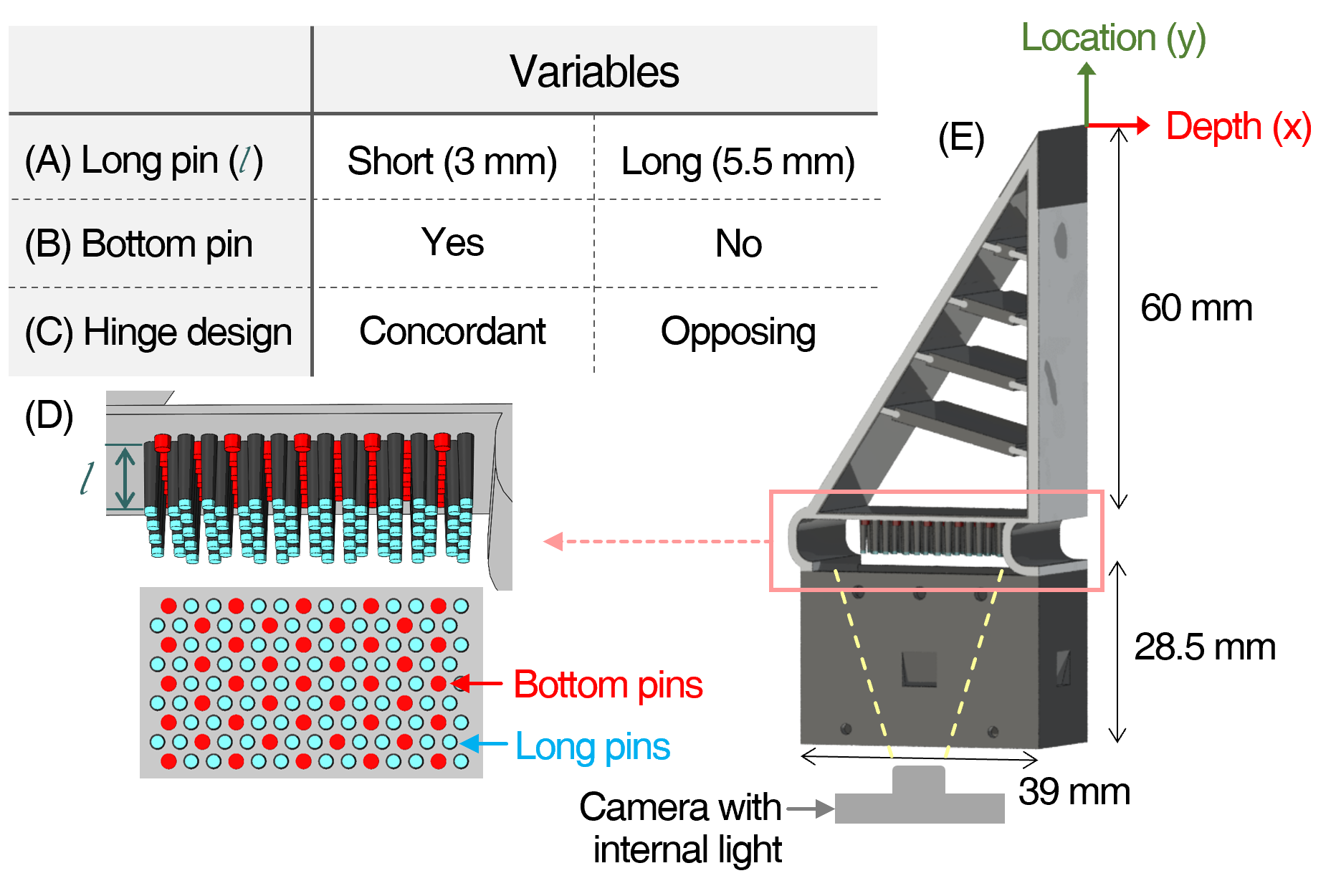}
      \vspace{-2ex}
      \caption{Key variables for effective sensing in the Fin-Ray finger and its rendered image. To enable effective sensing of the contact location and depth of an external object, we considered the following variables of the bottom crossbeam: (A) the length of the long pin, (B) the presence of bottom pins, and (C) the orientation of the two hinges. The sensing of location and depth is achieved through the translation and deformation of (D) the bottom crossbeam, while an internal camera of the Fin-Ray finger (E) captures the marker movements.}
      \label{fig:FinRay}
   \end{figure}

The Fin-Ray finger is designed to achieve sufficient flexibility under contact forces below 3\,N, enabling effective transmission of translation and deformation to the bottom crossbeam (Refer to the pink rectangle in Fig.~\ref{fig:FinRay}(E)). This flexibility is crucial for ensuring that the movement of the pin-marker array beneath the beam varies accordingly under diverse beam deformations, which is essential for accurate sensing.

To achieve this optimized flexibility, the finger was fabricated using a multimaterial 3D printer (Stratasys J826 Prime), which is capable of simultaneously printing materials with significantly different mechanical properties, such as the soft material Agilus30 and the rigid material Vero. The structure was printed with a total thickness of 1.5\,mm, maintaining a rigid-to-soft material thickness ratio of 2:1 (1.0\,mm outside to 0.5\,mm inside). This configuration was engineered to facilitate adequate deformation while ensuring structural integrity under pressing forces below 3\,N.

\subsubsection{Design Optimization}
For enhanced sensing performance in the Fin-Ray finger, it is essential to efficiently transfer the translation and deformation of the finger pad, induced by external contact, to the bottom crossbeam, which serves as the primary sensing unit. First, to ensure that the internal camera effectively captures the deformation of the bottom crossbeam through the pin arrangements, we considered the length of the long pin (Fig.~\ref{fig:FinRay}(A), indicated as blue dots in Fig.~\ref{fig:FinRay}(D)). Furthermore, we examined whether the presence of base pins (red dots in Fig.~\ref{fig:FinRay}(D)) improves sensing capabilities (Fig.~\ref{fig:FinRay}(B)). The pins have a diameter of 0.9~mm, with an inter-pin spacing of 1.7~mm. Finally, we investigated the effect of hinge orientation on the finger's translation behavior. Specifically, we analyzed whether orienting the two hinges to face each other (opposing configuration, as shown in Fig.~\ref{fig:Math}) helps suppress translation caused by contact, or whether aligning them in the same direction (concordant configuration, as shown in Fig.~\ref{fig:FinRay}(E)) induces greater translation, potentially enhancing sensing performance. The specific conditions for these comparisons are detailed in Fig.~\ref{fig:FinRay}(A-C) and the corresponding results are presented in Section~\ref{sub:Sensing_Opt}.
   
\subsection{Neural Network Training}
\begin{figure}[t]
      \centering
      \vspace{1ex}
      \includegraphics[width = 0.80\columnwidth]{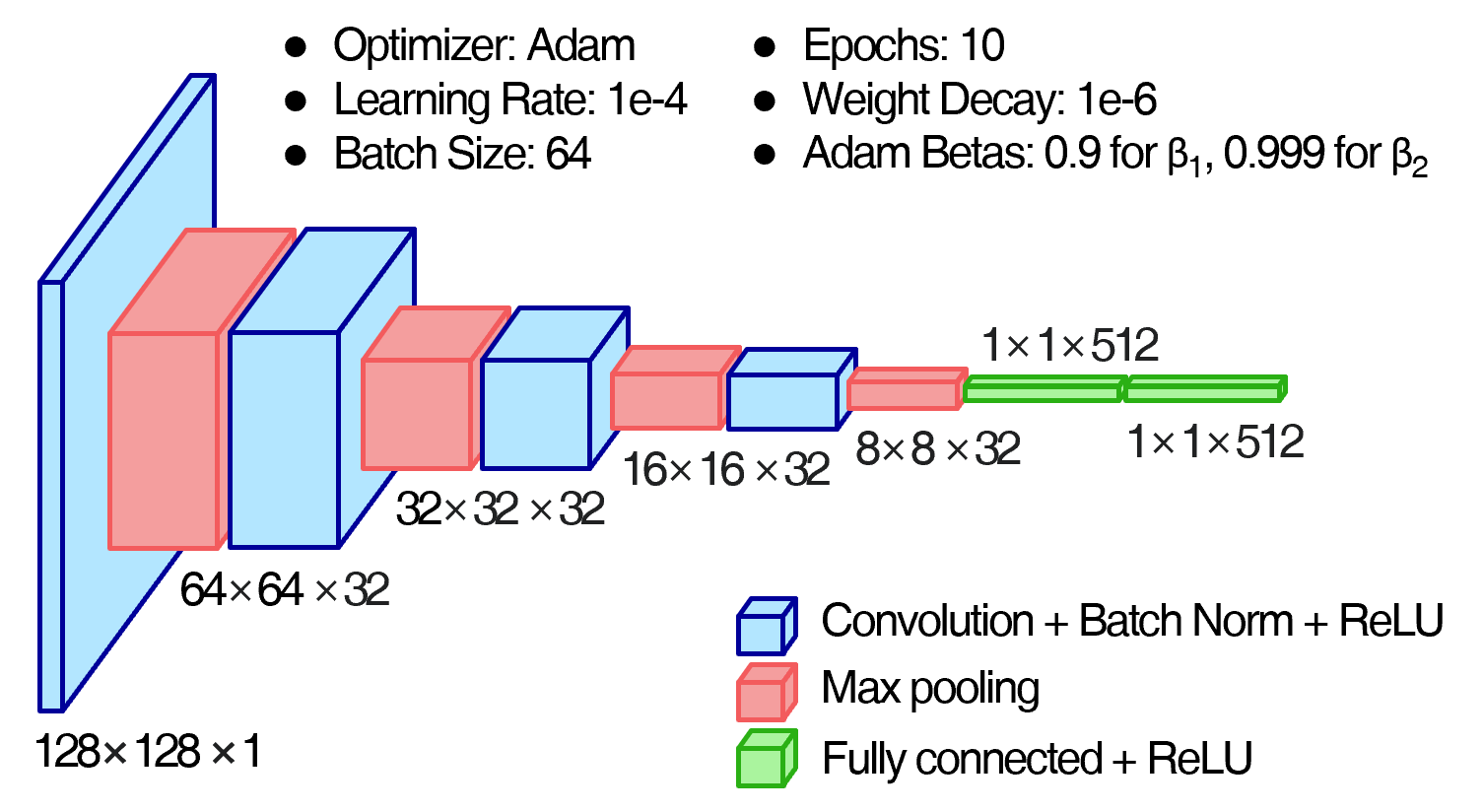}
      \vspace{-1ex}
      \caption{The deep neural network architecture inputs a 1-channel, 128$\times$128 image using four sequential convolutional blocks. Each block consists of a 2D convolution (32 filters; kernel sizes 11, 9, 7, or 5), batch normalization, a ReLU activation, and max-pooling. The 8$\times$8$\times$32 feature map is then flattened and passed through two 512-neuron fully-connected regression layers.}
      \label{fig:CNN}
   \end{figure}

The prediction of contact location and depth is achieved using a Convolutional Neural Network (CNN; Fig.~\ref{fig:CNN}) trained to associate the arrangement of pins beneath the bottom crossbeam with the two aforementioned contact conditions. To collect training data, the rigid rectangular pillar of the Fin-Ray finger was fixed to a horizontal mounting plate, upon which was also mounted a 4-axis desktop robot arm (Dobot MG400) that was equipped with an indenter to make contact with the Fin-Ray at various locations and depths (see Fig.~\ref{fig:Intro}). The internal camera of the Fin-Ray finger then captured tactile images of the marker-tipped pins following each contact event. The indentation depth varied from 1\,mm to 5.5\,mm from the finger surface, with contact location between 10\,mm and 50\,mm from the tip of the finger toward the bottom crossbeam.

To generate the dataset, 4,000 target 2D points (the pair of indentation depth and contact location) were randomly sampled within this range. For each sampled point, the indenter was actuated to induce deformation, and the internal camera acquired an image of the resulting pin array displacement. The collected dataset of 4,000 images was subsequently partitioned into training and evaluation sets, maintaining an 80:20 ratio.

Since only the movement of the marker-tipped pins is required for analysis, the captured tactile images were binarized, assigning 1 to the pin tips and 0 to all other areas, thereby reducing data complexity (see Fig.~\ref{fig:Comp}). In total, 4,000 pairs of 2D contact information (\textit{i.e.}, location and depth) tactile image datasets were generated, with 80\% for training and the remaining 20\% reserved for validation.

\subsection{Usefulness of Hinge Mechanism}
\label{sec:HingeMechanism}
\begin{figure}[t]
      \centering
      \includegraphics[width = \columnwidth]{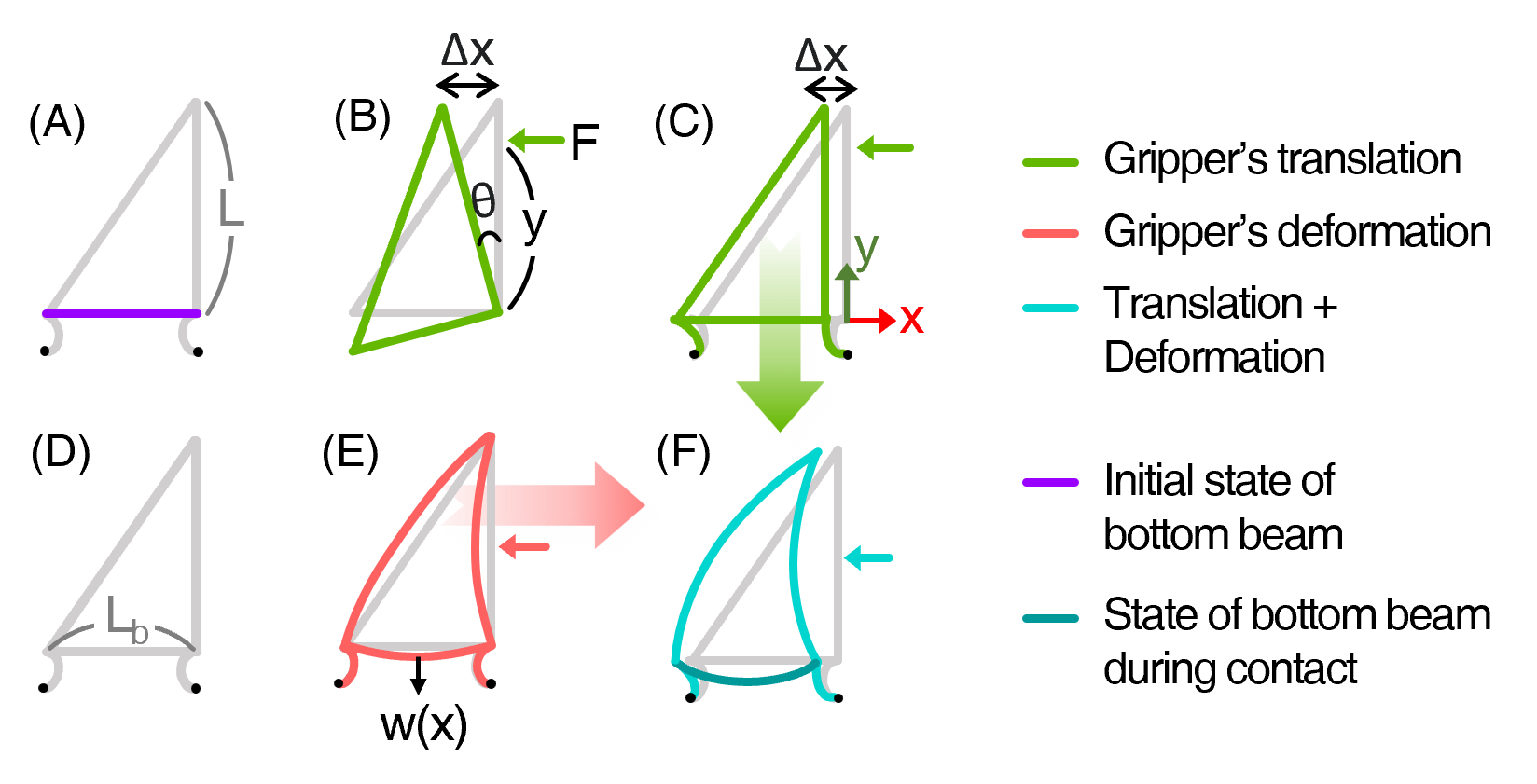}
      \vspace{-3ex}
      \caption{Analysis of the motion of the bottom beam in the Fin-Ray finger induced by the added hinges. Considering the $x$, $y$ coordinate system defined in (C), when a perpendicular force $F$ is applied to the finger at a location offset by $y$, the resulting deformation is a combination of two effects: (1)~the translation of the gripper due to the hinge mechanism (A–C) and (2)~the inherent deformation of the Fin-Ray finger (D, E). These combined effects ultimately lead to the deformed state illustrated in (F).}
      \label{fig:Math}
   \end{figure}

The primary advantage of our Fin-Ray finger is to simultaneously detect the contact location $y$ and the indentation depth $x$ (origin of the $xy$ coordinate system defined in Fig.~\ref{fig:Math}(C)). This dual-detection capability is made possible by the integration of an additional hinge mechanism located at the base of the Fin-Ray finger. In this section, we first mathematically demonstrate how these additional hinges enable the detection of both $x$ and $y$ contact information.

As depicted in Fig.~\ref{fig:Math}, we hypothesize that an object contacts the finger's finger pad at a specific location $y$, exerting a perpendicular force $F$, where $F$ is assumed to be directly proportional to the indentation depth $x$. Based on this hypothesis, the deformation of the bottom crossbeam (highlighted in purple in Fig.~\ref{fig:Math}(A)) can be analyzed through two primary deformation phenomena: (1) displacement in the $x$ direction (Fig.~\ref{fig:Math}(A-C), represented as $x' = x - \Delta x$) and (2) beam deflection, denoted as $w(x)$ in Fig.~\ref{fig:Math}(E).

We first perform a step-by-step mathematical analysis of each deformation component and subsequently integrate the two (Fig.~\ref{fig:Math}(F)) to understand how the hinge mechanism facilitates the detection of contact location in both the $x$ and $y$ directions when a force $F$ is applied at $y$. Ultimately, the objective is to derive an expression for the overall deformation of the bottom crossbeam, represented as $w(x', y, F)$.

\subsubsection{Shift Due to Applied Force}
When an external force $F$ is applied at a certain height $y$ , the resulting rotation angle, denoted as $\theta(y, F)$, can be expressed as $\theta(y, F) = -Fy/k$, where $k$ represents the torsional stiffness (Fig.~\ref{fig:Math}(B)). This rotation can be considered equivalent to an overall horizontal shift $\Delta x(y, F)$ of the finger if the rotation angle is sufficiently small (Fig.~\ref{fig:Math}(C)). Assuming that the finger has a length $L$, the shift in the $x$ direction can be approximated using the small-angle approximation ($\sin \theta \approx \theta$) as: $\Delta x(y, F) \approx L \sin \theta(y, F) \approx L \theta(y, F)$. Substituting $\theta(y, F)$ into this equation yields
    \begin{equation}
        \Delta x(y, F) = -\frac{L F y}{k}.
    \end{equation}
The new position of the bottom crossbeam, accounting for this shift, is therefore given by
    \begin{equation}
        x' = x - \Delta x(y, F) = x + \frac{L F y}{k}.
    \end{equation}
This result indicates that every point along the bottom beam experiences a horizontal displacement $\Delta x(y, F)$, which is directly proportional to both the magnitude of the applied force and its acting height.

\subsubsection{Deflection}
The bottom crossbeam also undergoes deflection due to the structural constraints of the finger deformed by the applied force $F$ (Fig.~\ref{fig:Math}(E)). This deflection behavior can be described using beam theory, where the deflection $w(x, y, F)$ follows from classical beam equations.

Assuming the bottom crossbeam of the Fin-Ray finger is a simple fixed-fixed beam, and considering an external force $F$ applied at a specific height $y$, the Euler-Bernoulli beam equation applies as follows
    \begin{equation}
        EI \frac{d^2 w}{d x^2} = M (y) = Fy,
    \end{equation}
where $E$ is the Young's modulus, $I$ is the second moment of inertia of the beam's cross-section, and $M(y)$ is the bending moment induced by the applied force at height $y$. Given that the beam is fixed at both ends, the boundary conditions are $w(x=0) = w(x=L_b) = 0$, where $L_b$ represents the length of the bottom crossbeam (Fig.~\ref{fig:Math}(D)).

By integrating the Euler-Bernoulli equation twice and applying the boundary conditions, an expression for the deflection of the beam is obtained:
\begin{equation}
    w(x, y, F) = \frac{F x y (x - L_b)}{2EI},
\end{equation}
which describes the deflection profile of the bottom crossbeam under the applied force $F$, showing that the deflection is influenced by the force magnitude, the height at which it is applied, and the beam's material and geometric properties.

    \begin{figure}[t]
      \centering
      \includegraphics[width = 0.85\columnwidth]{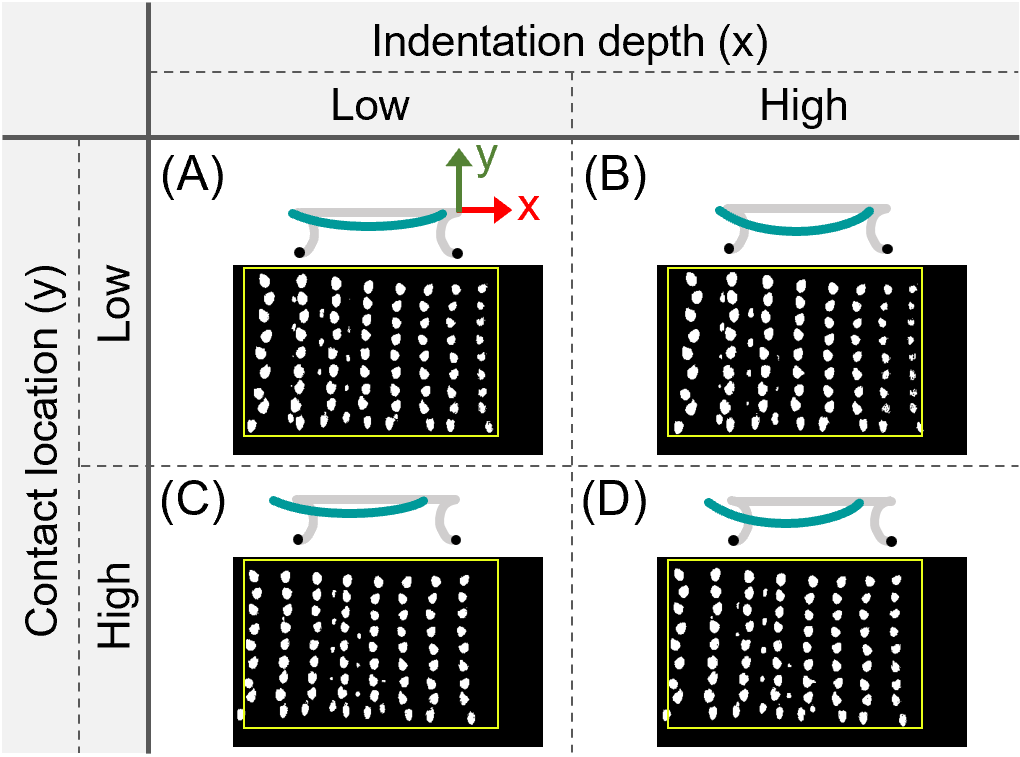}
      \caption{An illustration of the translation and deformation of the bottom crossbeam in response to the indentation depth $x$ and contact location $y$ of an object in contact with the Fin-Ray finger, along with the corresponding experimentally measured tactile images, which have been binarized from the original data. To facilitate comparison of marker movements under the four conditions, a yellow box has been placed at a fixed reference position.}
      \label{fig:Comp}
   \end{figure}

\subsubsection{Overall Deformation}

Considering the shift of the beam, we substitute $x$ with the shifted coordinate $x'$:
    \begin{align}
        w(x', y, F) &= \frac{F (x - \Delta x) y (x - \Delta x - L_b)}{2EI}\\
        &= \frac{F y}{2EI} \left( x + \frac{L F y}{k} \right) \left( x + \frac{L F y}{k} - L_b \right)\label{eq:overall}.
    \end{align}

This equation accounts for both the translational shift and deflection of the bottom crossbeam given the contact location $y$ and the applied force $F$, which is proportional to the indentation depth $x$ (illustrated by the blue-green colored beam in Fig.~\ref{fig:Math}(F)). Therefore, the integration of the hinge mechanism enables the dual-detection capabilities in the Fin-Ray finger, allowing it to simultaneously identify both the contact location and the indentation depth.

\subsection{Deformation of the Marker-tipped Pins}
In the above section, we derived Eq.~(\ref{eq:overall}) that generates discernible deformations for various contacts at different coordinates $x$ and $y$. However, directly applying this equation to inversely estimate the 2D contact information ($x, y$) is practically challenging. First, it is difficult to accurately identify the Young's modulus of the composite material used in the finger, and also the effects of the viscosity of the material. Furthermore, factors such as the thickness of the beam and variations in deformation caused by structural constraints of the Fin-Ray finger with contact location have not been considered.

Therefore, we adopted a data-driven approach, and used the mathematical approach above to understand the underlying mechanism. Using the marker-tipped pins attached underneath the bottom crossbeam (Fig.~\ref{fig:FinRay}(E)), we capture the deformation patterns represented in the tactile images. These images, which vary according to deformation, are then processed using neural network techniques to estimate the contact depth and location. This approach enables the effective reconstruction of the 2D contact information $(x,y)$ without the need for precise material property characterization or complex mechanical modeling.

The sensing principle of the Fin-Ray finger with the hinge mechanism can be explained in relation to the behavior of the bottom crossbeam, which can be categorized into two main effects: translation and deformation. 

First, as the indentation depth increases, the overall deformation of the Fin-Ray finger becomes more pronounced, resulting in greater deflection of the bottom crossbeam (left versus right columns in Fig.~\ref{fig:Comp}). Second, as the contact location shifts from the lower region near the hinge to the upper tip of the finger, the moment induces a larger translational displacement (upper versus lower rows in Fig.~\ref{fig:Comp}), even though the applied force remains constant.

Figure~\ref{fig:Comp} presents the analytically modeled movement of the bottom crossbeam under four extreme conditions, along with processed images of the pin-marker array from the fabricated Fin-Ray finger. As the indentation depth increases, the marker arrangement expands due to the greater deformation of the bottom crossbeam. Additionally, as the contact location shifts from a lower to a higher position, the entire array exhibits a more pronounced leftward shift.


\section{Results}
\label{sec:results}
This results section has two main parts. First, we compare the sensing performance of three design variables to determine the optimal Fin-Ray finger configuration for maximizing contact depth and location sensing accuracy (Sec.~\ref{sub:Sensing_Opt}). Second, we present two experiments conducted with the optimized finger design. The first experiment investigates the effect of indenter shape on sensing accuracy by comparing the Mean Absolute Error (MAE) of the predictions (Sec.~\ref{sub:Indenter_Tip}). The second experiment assesses the finger’s 2D sensing capability in a pick-and-place task with a cylinder, demonstrating the practical utility of its tactile sensing (Sec.~\ref{sec:PosErr}).

\subsection{Design Optimization for Better Sensing}
\label{sub:Sensing_Opt}
\begin{figure}[t]
      \centering
      \includegraphics[width = 1.0\columnwidth]{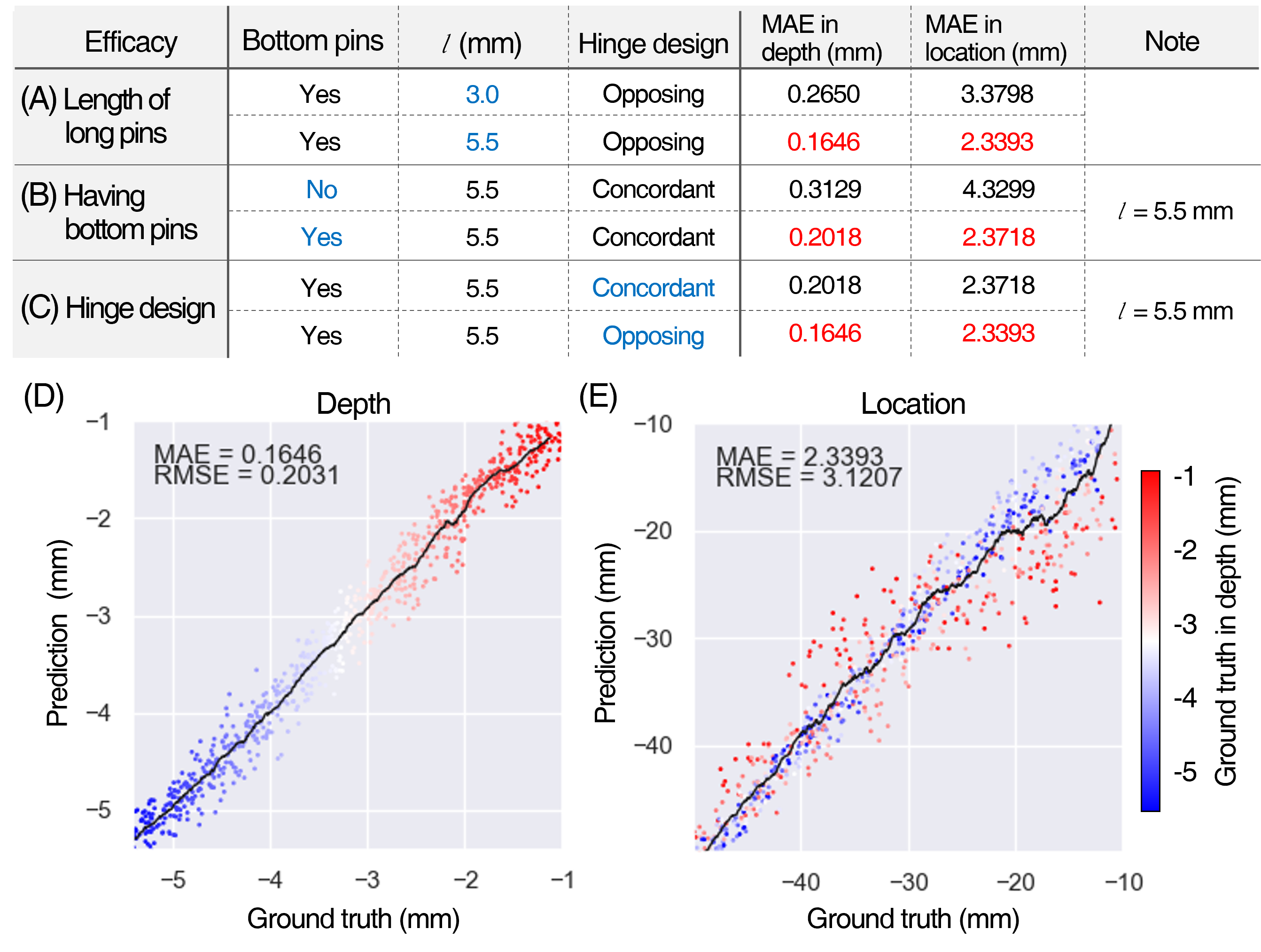}
      \caption{Results for various pin configurations and the presence or absence of a hinge (A–C), and example plots comparing predicted and ground-truth depth and contact location with the best parameters (D, E; blue–white-red colors indicate the ground truth for depth).}
      \label{fig:FinRayOpt}
  \end{figure}

To maximize the sensing capability of the Fin-Ray finger, we evaluated its sensing performance based on the variables defined in Fig.~\ref{fig:FinRayOpt}(A–C). Performance was assessed by comparing the location and depth values predicted by the trained 2D CNN (20\% of dataset) with the ground truth values. Fig.~\ref{fig:FinRayOpt}(A–C) summarizes the MAE values for each condition.

\subsubsection{Effect of Pin Length}
First, longer pins demonstrated superior sensing performance (Fig. ~\ref{fig:FinRayOpt}(A)). When comparing fingers with pin lengths of 3\,mm and 5.5\,mm, both depth and location sensing showed lower MAE values for the finger with 5.5\,mm pins. With 5.5\,mm pins, the MAE was 0.17\,mm for depth sensing and 2.33\,mm for location sensing.

\subsubsection{Effect of Base Pins}
Second, the presence of a secondary layer base (shorter) pins was found to improve sensing capability. As shown in Fig.~\ref{fig:FinRayOpt}(B), when comparing MAE values while keeping all other variables fixed, the finger with base pins exhibited lower MAE in both depth and location sensing. Specifically, the addition of base pins to the existing 5.5\,mm pins reduced the MAE by 0.11\,mm (2.5\% decrease in error rate) for depth sensing and by 1.96\,mm (4.9\% decrease in error rate) for location sensing.

\subsubsection{Effect of Hinge Orientation}
Third, hinge orientation was found to significantly impact the sensing sensitivity of the Fin-Ray finger (Fig.~\ref{fig:FinRayOpt}(C)). The opposing hinge design exhibited higher sensitivity compared to the concordant hinge pair. When 5.5\,mm pins and base pins were used together, and the hinge orientation was opposing, both depth and location sensing achieved the lowest MAE.

\subsubsection{Validation with the Optimal Variable Combination}
The optimal combination of variables that resulted in the lowest MAE was found in the Fin-Ray finger with a long pin length of 5.5\,mm, the presence of base pins, and an opposing hinge design. The validation results for depth and location sensing using this finger configuration are plotted in Fig.~\ref{fig:FinRayOpt}(D, E). The MAE for depth sensing was 0.16\,mm, which corresponds to an error rate of 3.7\% over a depth-sensing range of 4.5\,mm. Meanwhile, the error rate for location sensing was 5.9\%.

\subsection{Online Experiments with the Optimized Gripper}
Using the best-performing finger design identified above, which featured long pins with base pins and opposing hinges, along with the trained CNN model, a series of tests were conducted to evaluate the robustness of the sensorised finger and its suitability for practical applications.

\subsubsection{Robustness of Contact Localization across Indenters}
\label{sub:Indenter_Tip}
\begin{figure}[t]
      \centering
      \includegraphics[width = \columnwidth]{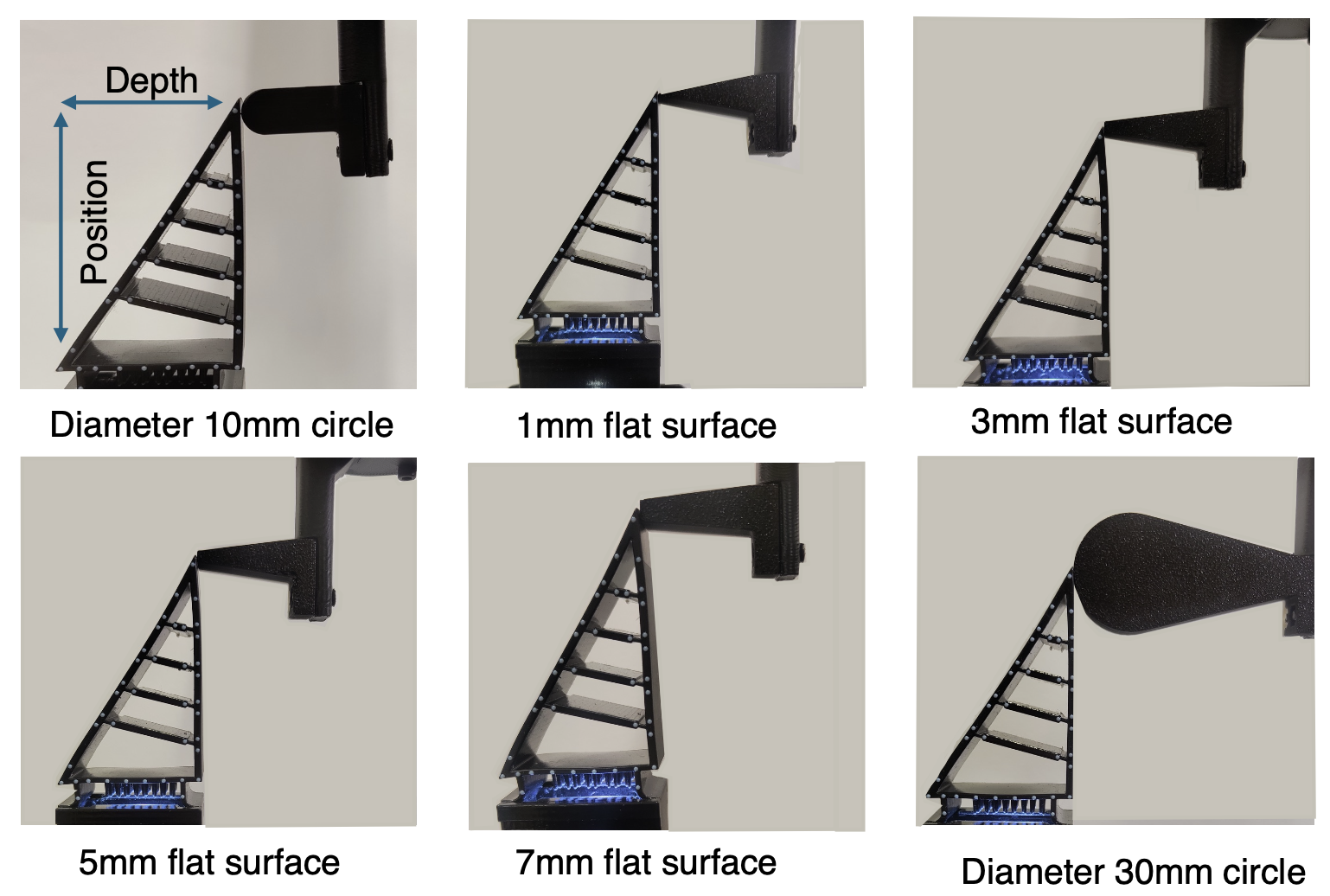}
      \caption{Six indenter shapes used in the experiments. The first was used for training data collection, and the remaining five for testing.}
      \label{different indenters}
  \end{figure}

The CNN model for contact localization was trained using one indenter shape, which was a circular indenter with 10\,mm diameter (first subfigure in Fig.~\ref{different indenters}). To assess that the sensing capability generalizes across indenters of varying shape and sizes, an experiment was conducted to validate the model's robustness in recognizing contact locations from objects with larger dimensions and different geometries. Figure \ref{different indenters} shows six different contact scenarios, including the indenter used to train the CNN model (first subfigure), and the other subfigures feature flat-surfaces with heights of 1\,mm, 2\,mm, 3\,mm, 5\,mm, and 7\,mm, as well as a circular indenter with a 30\,mm diameter, matching the diameter of the cylindrical pillar used in Sec.~\ref{sec:PosErr}). These experiments are designed to simulate non-precise, distributed contact conditions, where multiple contact points are present. The objective is to evaluate the sensor's accuracy in estimating the central axis of contact, despite the presence of extended or multi-point contact regions.

Using the data collection setup (Fig. 1), the indenters were systematically replaced, then 30 contact trials onto the fixed finger were performed for each indenter type. These trials were conducted at randomized positions within 1\,mm intervals, covering an indentation depth range of 1--5\,mm and a contact location range of 5--55\,mm. The sensing errors in depth and location were calculated by comparing the ground-truth contact positions of the indenter's tip from the robot inverse kinematics with the corresponding predictions generated by the trained CNN model.

In Table~\ref{figure experiment1}, the 3\,mm flat indenter demonstrated the most balanced sensing performance overall when considering both depth and location errors, although it did not achieve the lowest error in each individual metric. The lowest depth error was recorded with the 30\,mm diameter circular indenter, while the lowest location error was observed with the 3\,mm flat indenter. Notably, the indenter with the lowest mean error also had the smallest standard deviation, indicating consistent sensing performance.

  \begin{table}[t]
      \centering
      \caption{Experiment 1: sensing performance for six indenters}
      \label{figure experiment1}
      \includegraphics[width = 0.8\columnwidth]{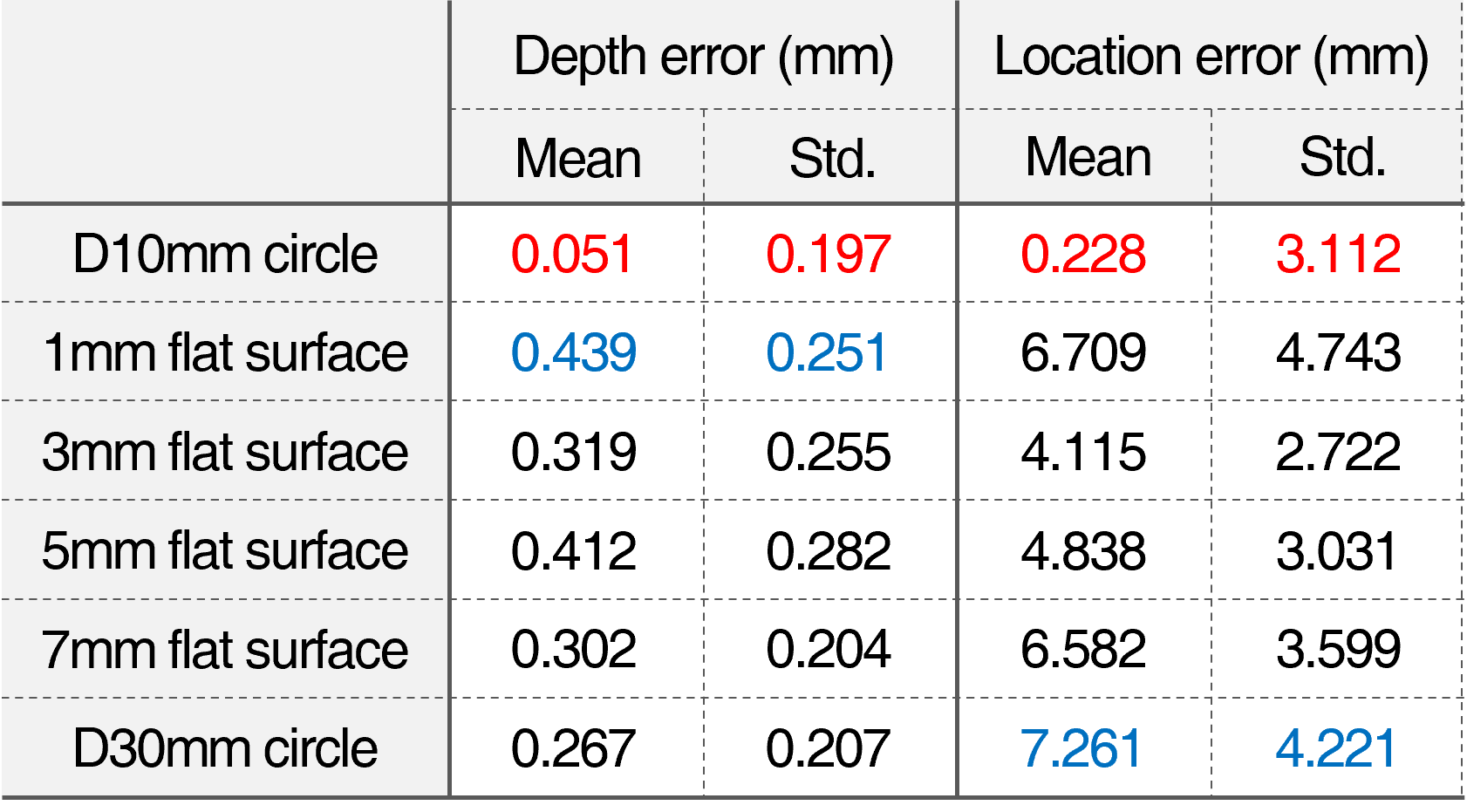}
  \end{table}

To assess whether the variations in sensing errors were statistically significant across different indenters, we conducted a pairwise Tukey's HSD test. No indenter pair showed a statistically-significant difference (p \textless 0.05) in either depth or location errors. This result supports the validity of the dataset collected using a single indenter type (the 10\,mm diameter circular indenter used during training) despite minor differences in error magnitude across other indenter shapes.

\subsubsection{Demonstration of Positional Error Correction}
\label{sec:PosErr}
\begin{figure}[t]
      \vspace{-2ex}
      \centering
      \includegraphics[width = \columnwidth]{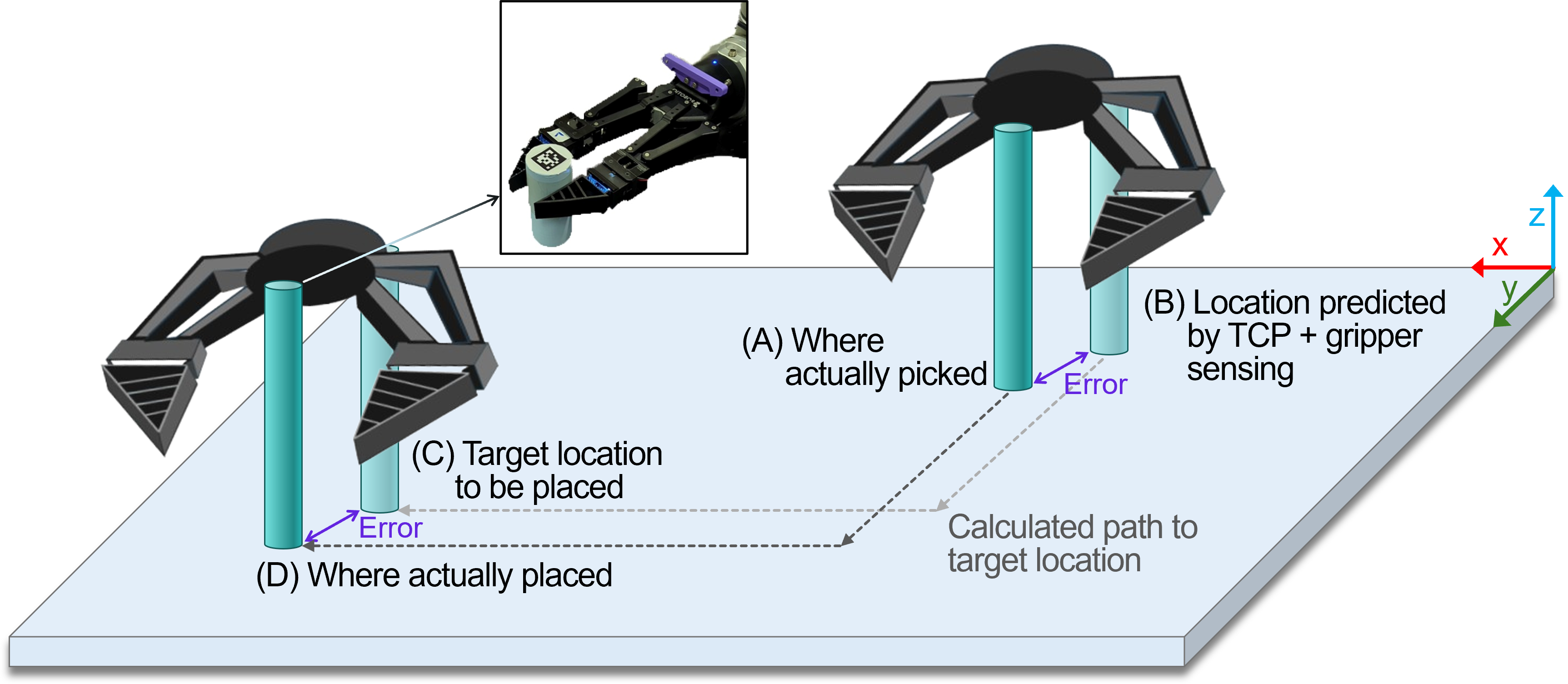}
      \caption{Pick-and-place task utilizing the sensing capabilities of the tactile Fin-Ray gripper. (A) The robot picks a randomly-placed cylinder within the gripper’s workspace, then (C) places it at a predefined target location. (B) Due to sensing errors, the estimate of the cylinder's position, (D) results in a final placement offset from the target location. The placement error is calculated as the Euclidean distance between the points in (C) and (D), using the position information from an ArUco marker on the cylinder top.}
      \label{figure experiment2}
  \end{figure}

As a use-case for the sensorized Fin-Ray finger, we designed a demonstration to perform precise pick-and-place tasks under conditions of uncertain picking position. A pair of sensorized Fin-Ray fingers was mounted onto a Robotiq 2F-140 servo gripper (see Fig.~\ref{figure experiment2}), which was attached to a Universal Robots UR-10 robot arm. The tool center point (TCP) of the robot was defined as the midpoint between the two Fin-Ray fingers when they are in the closed position.

Specifically,  the cylinder's location is estimated by combining the TCP information with the contact depth and location information obtained through tactile sensing within the gripper's workspace (Fig.~\ref{figure experiment2}(B)). The distance between the actual location and predicted location of the cylinder is defined as the prediction error (shown as ``Error'' in Fig.~\ref{figure experiment2}). A larger prediction error denotes a greater deviation of the placed cylinder from the intended target location (Fig.~\ref{figure experiment2}(C)).

The prediction error directly translates to the final placement error, which is calculated as the Euclidean distance between the target location and the actual final position of the cylinder (Figs~\ref{figure experiment2}(C,D)). To accurately determine the actual position of the cylinder, an ArUco marker was attached to the top surface of the cylinder (see Fig.~\ref{figure experiment2}), and the $(x, y)$ coordinates extracted from the captured image.

\begin{figure}[t]
      \centering
    \vspace{1ex}
      \includegraphics[width = 0.95\columnwidth]{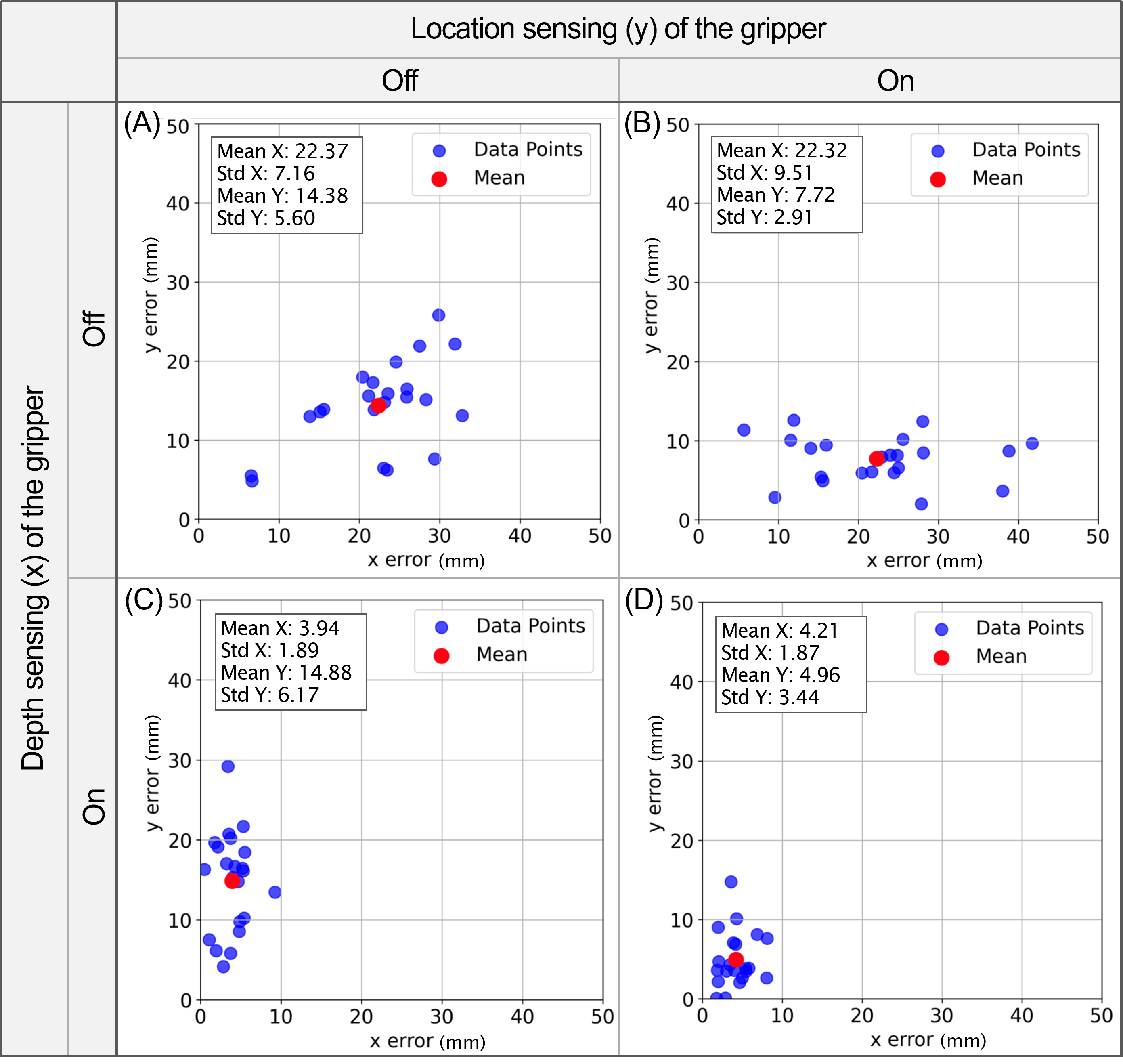}
      \caption{Scatter plots of placement errors in the pick-and-place task under different tactile sensing conditions of the Fin-Ray finger. Condition (D), where both depth and location sensing capabilities were enabled, exhibited the smallest error distribution, while condition (A), with no sensing capabilities used, resulted in the widest error spread.}
      \label{pick_and_place_task}
  \end{figure}

The pick-and-place task was conducted under four different sensing conditions to evaluate the impact of tactile sensing on placement accuracy. In the first condition, the Fin-Ray finger operated (1) without utilizing any sensing capabilities. In the second, (2) only the depth sensing function was used, with the contact location fixed at the center of the gripper. In the third condition, (3) only the contact location sensing along the length of the Fin-Ray finger was employed, while the cylinder was assumed to be centered during placement. In the final condition, (4) both depth and location sensing capabilities were used to precisely predict the placement location. 20 trials were conducted for each scenario, and the results are shown in Fig.~\ref{pick_and_place_task}. It is important to note that in all four conditions, the robot was controlled to ensure that the cylinder remained stationary during the sensing phase of the grasping process. 

Fig.~\ref{pick_and_place_task} presents the scatter plots of placement errors obtained from 20 repeated pick-and-place tasks conducted under each of the four sensing conditions. When neither depth nor location sensing capabilities of the Fin-Ray finger were utilized (Fig.~\ref{pick_and_place_task}(A)), large errors were observed along both the x and y axes. In contrast, when the sensing functions were enabled, both the mean and standard deviation of the errors were significantly reduced. Notably, the smallest error distribution was achieved when both depth and location sensing were employed (Fig.~\ref{pick_and_place_task}(D)).

\section{Discussion}
\subsection{Challenges in Analytical Modeling}
Accurately predicting the deformation and translation of the bottom crossbeam in response to external contact is challenging. Although Eq.~\ref{eq:overall} provides a mathematical representation of the bottom crossbeam's movement, determining precise values for Young’s modulus ($E$) and the second moment of area ($I$) remains difficult. Specifically, even minor changes in the number, position, or thickness of the Fin-Ray finger’s crossbeams would require a complete re-optimization of these parameters. Furthermore, even if accurate values for $E$ and $I$ are obtained, practical implementation would still require a sensing system (\textit{e.g.}, a camera) to measure beam movement in real-world applications.

Given these constraints, a CNN gives the most suitable approach. By representing the movement (deformation and translation) of the bottom crossbeam as changes in the marker-pin array arrangement, this method enables easy sensing via a camera. Moreover, this data-driven approach has the potential for seamless adaptation across various soft robotics applications.
\vspace{-1ex}
\subsection{Design Insights for Enhancing Sensing Performance}
Achieving high sensing performance in the Fin-Ray finger requires a hybrid pin configuration consisting of long and base pins, and an opposing hinge orientation (Fig.~\ref{fig:FinRayOpt}(A–C)). 5.5\,mm long pins outperformed shorter 3.0\,mm pins in sensing accuracy (Fig.~\ref{fig:FinRayOpt}(A)). This is because longer pins amplify the deformation of the bottom crossbeam, allowing the CNN to more effectively distinguish subtle variations.




Due to their shorter length, only bottom pins near the optical axis remain visible, while those farther away are occluded by the longer pins. As a result, the visibility of base markers changes with the translation of the crossbeam, providing spatial information (Fig.~\ref{fig:Comp}(A), (C)).


Lastly, the opposing hinge configuration provides stiffness to balance translation and deformation. In the concordant hinge configuration, small contact forces cause large translations, limiting overall deformation. In contrast, the opposing hinge design increases resistance to translation, resulting in more pronounced deformation and enhancing sensing accuracy.

\vspace{-1ex}
\subsection{Designs for Enhanced Sensing Performance}
The performance of our Fin-Ray finger is closely linked to the optimization of its flexibility and structural design. Physical contact with an external object induces both deformation and translation of the finger, reflected in the combined marker pattern of the bottom crossbeam. Optimizing either the mechanical structure or the material softness to maximize movement variation under specific contact conditions could significantly enhance the overall sensing performance.

In our current design, when sufficient contact depth is applied (blue dots in Fig.~\ref{fig:FinRayOpt}(D)), the finger also exhibits higher accuracy in location sensing (blue dots in Fig.~\ref{fig:FinRayOpt}(E)). This suggests a strong correlation between depth-induced deformation and location sensing performance. If a hinge structure could be designed to induce greater translation under shallow contact depths, while limiting translation under deep contact conditions, it could potentially lead to improved accuracy in location sensing by increasing the range and resolution of detectable motion patterns.

\vspace{-1ex}
\subsection{Limitations and Future Work}
A Fin-Ray gripper's passive conformation typically creates a single curvature, making multi-point sensing inherently difficult. Acknowledging this single-contact constraint, we tested the system's performance using simple indenters (Table~\ref{figure experiment1}) to probe responses to geometric variations and generalization to a larger size. A comprehensive assessment using complex object shapes is essential future work.

This multi-point challenge is linked to resolving force distribution. In Fin-Ray grippers, force is proportional to deformation; our monolithic structure is thus limited to inferring a single, aggregate force value.

However, a design with multiple, thin Fin-Ray layers could exhibit complex deformations, suggesting a pathway to resolving force distribution. This layered approach could also aid in grasping asymmetric objects, as recently demonstrated in work on Fin-A-Rays~\cite{Lee25_FinARay_SoRo}.

Another limitation is indirect sensing. Direct approaches, like the GelSight Fin Ray~\cite{Wang24_GelSightFinray}, utilize high spatial resolution. In contrast, TacFinRay's indirect method acquires limited information (e.g. depth and location). Hence, high-level tasks like texture recognition require a direct sensing modality, although there are many tasks for which just the contact forces and positions are suffficient.

\section{Conclusion}
\label{sec:conclusion}
In this study, we proposed a novel sensorized Fin-Ray gripper that enables indirect tactile sensing by capturing deformation patterns of a bottom crossbeam. Through the integration of a hinge mechanism between the compliant gripper structure and a rigid camera module, the gripper achieved simultaneous measurement of contact location and indentation depth. By processing marker/pin array deformations with a convolutional neural network, the system eliminated the need for complex mechanical modeling. Optimization of pin length, bottom pin configuration, and hinge orientation significantly improved sensing accuracy. Validation experiments demonstrated robust generalization across various contact geometries, and a pick-and-place task under uncertain conditions confirmed the practical utility of the proposed tactile sensing approach. Overall, the presented methodology offers a lightweight and scalable solution and lays the groundwork for future enhancements in indirect tactile sensing for soft robotic applications.

\section*{Acknowledgments}
The authors thank Ocado Technology for providing a robot arm (UR-10) to conduct the object manipulation experiments. 





\bibliographystyle{unsrt}
\bibliography{./bib/paper}

@Article{Sferrazza19Sensors_OTS,
AUTHOR = {Sferrazza, Carmelo and D’Andrea, Raffaello},
TITLE = {Design, Motivation and Evaluation of a Full-Resolution Optical Tactile Sensor},
JOURNAL = {Sensors},
VOLUME = {19},
YEAR = {2019},
NUMBER = {4},
ARTICLE-NUMBER = {928},
PubMedID = {30813292},
ISSN = {1424-8220}}

@INPROCEEDINGS{Yuan15ICRA_SlipShear,
  author={Yuan, Wenzhen and Li, Rui and Srinivasan, Mandayam A. and Adelson, Edward H.},
  booktitle={2015 ICRA}, 
  title={Measurement of shear and slip with a GelSight tactile sensor}, 
  year={2015},
  volume={},
  number={},
  pages={304-311}
}

@INPROCEEDINGS{Fishel12BioRob_Biotac,
  author={Fishel, Jeremy A. and Loeb, Gerald E.},
  booktitle={BioRob}, 
  title={Sensing tactile microvibrations with the BioTac — Comparison with human sensitivity}, 
  year={2012},
  volume={},
  number={},
  pages={1122-1127}
}

@ARTICLE{Bhirangi23RAL_tactile_sensing,
  author={Bhirangi, Raunaq and DeFranco, Abigail and Adkins, Jacob and Majidi, Carmel and Gupta, Abhinav and Hellebrekers, Tess and Kumar, Vikash},
  journal={IEEE RA-L}, 
  title={All the Feels: A Dexterous Hand With Large-Area Tactile Sensing}, 
  year={2023},
  volume={8},
  number={12},
  pages={8311-8318}}

@article{Lepora24SciRobotics_RobotHand,
author = {Nathan F. Lepora },
title = {The future lies in a pair of tactile hands},
journal = {Sci. Robot.},
volume = {9},
number = {91},
pages = {eadq1501},
year = {2024}}

@ARTICLE{Lepora21Sensors_TacTipReview,
  author={Lepora, Nathan F.},
  journal={IEEE Sens. J.}, 
  title={Soft Biomimetic Optical Tactile Sensing With the TacTip: A Review}, 
  year={2021},
  volume={21},
  number={19},
  pages={21131-21143}
}

@ARTICLE{Lu22RAL_FourFinger,
  author={Lu, Zeyu and Guo, Haotian and Zhang, Wensi and Yu, Haoyong},
  journal={IEEE RA-L}, 
  title={GTac-Gripper: A Reconfigurable Under-Actuated Four-Fingered Robotic Gripper With Tactile Sensing}, 
  year={2022},
  volume={7},
  number={3},
  pages={7232-7239}
}

@article{Guo24AIS_FinRayARUCO,
author = {Guo, Ning and Han, Xudong and Zhong, Shuqiao and Zhou, Zhiyuan and Lin, Jian and Wan, Fang and Song, Chaoyang},
title = {Reconstructing Soft Robotic Touch via In-Finger Vision},
journal = {Adv. Intell. Syst.},
volume = {6},
number = {10},
pages = {2400022},
year = {2024}
}

@ARTICLE{Guo24TRO_StateEstimate,
  author={Guo, Ning and Han, Xudong and Zhong, Shuqiao and Zhou, Zhiyuan and Lin, Jian and Dai, Jian S. and Wan, Fang and Song, Chaoyang},
  journal={IEEE T-RO}, 
  title={Proprioceptive State Estimation for Amphibious Tactile Sensing}, 
  year={2024},
  volume={40},
  number={},
  pages={4684-4698}
}

@ARTICLE{Lu24TMech_DexiTac,
  author={Lu, Chenghua and Tang, Kailuan and Yang, Max and Yue, Tianqi and Li, Haoran and Lepora, Nathan F.},
  journal={IEEE/ASME Trans. Mechatron.}, 
  title={DexiTac: Soft Dexterous Tactile Gripping}, 
  year={2024},
  volume={},
  number={},
  pages={1-12}}

@ARTICLE{Xu21TRO_FinRayForce,
  author={Xu, Wenfu and Zhang, Heng and Yuan, Han and Liang, Bin},
  journal={IEEE TR-O}, 
  title={A Compliant Adaptive Gripper and Its Intrinsic Force Sensing Method}, 
  year={2021},
  volume={37},
  number={5},
  pages={1584-1603}
}

@article{Shan20IJRR_FinRayEffect,
author = {Xiaowei Shan and Lionel Birglen},
title ={Modeling and analysis of soft robotic fingers using the fin ray effect},
journal = {Int. J. Robot. Res.},
volume = {39},
number = {14},
pages = {1686-1705},
year = {2020}
}

@article{Yao24MMT_FinRayEffect2,
title = {Design optimization of soft robotic fingers biologically inspired by the fin ray effect with intrinsic force sensing},
journal = {Mech. Mach. Theory},
volume = {191},
pages = {105472},
year = {2024},
issn = {0094-114X},
author = {Jiaqiang Yao and Yuefa Fang and Xinhua Yang and Peiyi Wang and Luquan Li}
}

@ARTICLE{Chen23TRO_DeflectionSensing,
  author={Chen, Genliang and Tang, Shujie and Xu, Shaoqiu and Guan, Tong and Xun, Yuanhao and Zhang, Zhuang and Wang, Hao and Lin, Zhongqin},
  journal={IEEE TR-O}, 
  title={Intrinsic Contact Sensing and Object Perception of an Adaptive Fin-Ray Gripper Integrating Compact Deflection Sensors}, 
  year={2023},
  volume={39},
  number={6},
  pages={4482-4499}}

@misc{Wang24_GelSightFinray,
      title={GelSight FlexiRay: Breaking Planar Limits by Harnessing Large Deformations for Flexible,Full-Coverage Multimodal Sensing}, 
      author={Yanzhe Wang and Hao Wu and Haotian Guo and Huixu Dong},
      year={2024},
      eprint={2411.18979},
      archivePrefix={arXiv},
      primaryClass={cs.RO}
}

@ARTICLE{Li24_BioTacTip_RAL,
  author={Li, Haoran and Nam, Saekwang and Lu, Zhenyu and Yang, Chenguang and Psomopoulou, Efi and Lepora, Nathan F.},
  journal={IEEE RA-L}, 
  title={BioTacTip: A Soft Biomimetic Optical Tactile Sensor for Efficient 3D Contact Localization and 3D Force Estimation}, 
  year={2024},
  volume={9},
  number={6},
  pages={5314-5321}}

@article{Lloyd24TacServoing_IJRR,
author = {John Lloyd and Nathan F. Lepora},
title ={Pose-and-shear-based tactile servoing},
journal = {Int. J. Robot. Res.},
volume = {43},
number = {7},
pages = {1024-1055},
year = {2024}
}

@ARTICLE{Lambeta20DIGIT_RAL,
  author={Lambeta, Mike and Chou, Po-Wei and Tian, Stephen and Yang, Brian and Maloon, Benjamin and Most, Victoria Rose and Stroud, Dave and Santos, Raymond and Byagowi, Ahmad and Kammerer, Gregg and Jayaraman, Dinesh and Calandra, Roberto},
  journal={IEEE RA-L}, 
  title={DIGIT: A Novel Design for a Low-Cost Compact High-Resolution Tactile Sensor With Application to In-Hand Manipulation}, 
  year={2020},
  volume={5},
  number={3},
  pages={3838-3845}
}

@INPROCEEDINGS{Liu23RoboSoft_GelsighBabayFinray,
  author={Liu, Sandra Q. and Ma, Yuxiang and Adelson, Edward H.},
  booktitle={RoboSoft}, 
  title={GelSight Baby Fin Ray: A Compact, Compliant, Flexible Finger with High-Resolution Tactile Sensing}, 
  year={2023},
  volume={},
  number={},
  pages={1-8}
}

@misc{Zhang25_PneuGelSight,
      title={PneuGelSight: Soft Robotic Vision-Based Proprioception and Tactile Sensing}, 
      author={Ruohan Zhang and Uksang Yoo and Yichen Li and Arpit Argawal and Wenzhen Yuan},
      year={2025},
      eprint={2508.18443},
      archivePrefix={arXiv},
      primaryClass={cs.RO},
      url={https://arxiv.org/abs/2508.18443}, 
}

@article{Lee25_FinARay_SoRo,
author = {Lee, Loong Yi and Terrile, Silvia and Nam, Saekwang and Liang, Tianhao and Lepora, Nathan and Rossiter, Jonathan},
title = {Fin-A-Rays: Expanding Soft Gripper Compliance via Discrete Arrays of Flexible Structures},
journal = {Soft Robot.},
volume = {},
number = {},
pages = {},
year = {2025}
}

@article{Faris23_SoRo,
author = {Faris, Omar and Muthusamy, Rajkumar and Renda, Federico and Hussain, Irfan and Gan, Dongming and Seneviratne, Lakmal and Zweiri, Yahya},
title = {Proprioception and Exteroception of a Soft Robotic Finger Using Neuromorphic Vision-Based Sensing},
journal = {Soft Robot.},
volume = {10},
number = {3},
pages = {467-481},
year = {2023}
}

@ARTICLE{Susini25_RAL,
  author={Susini, Paolo and Pagnanelli, Giulia and Nam, Saekwang and Lepora, Nathan F. and Bianchi, Matteo},
  journal={IEEE RA-L}, 
  title={Data-Driven Compliance Discrimination via Biomimetic Soft Optical Tactile Sensors: Implementation and Benchmarking With a Model-Based Approach}, 
  year={2025},
  volume={10},
  number={8},
  pages={8083-8090}
}

@ARTICLE{Zhang24_Tacpalm,
  author={Zhang, Xuyang and Yang, Tianqi and Zhang, Dandan and Lepora, Nathan F.},
  journal={IEEE Sens. J.}, 
  title={TacPalm: A Soft Gripper With a Biomimetic Optical Tactile Palm for Stable Precise Grasping}, 
  year={2024},
  volume={24},
  number={22},
  pages={38402-38416}
}

@article{zhang2025soft,
  title={Soft robotic hand with tactile palm-finger coordination},
  author={Zhang, Ningbin and Ren, Jieji and Dong, Yueshi and Yang, Xinyu and Bian, Rong and Li, Jinhao and Gu, Guoying and Zhu, Xiangyang},
  journal={Nat. Commun.},
  volume={16},
  number={1},
  pages={2395},
  year={2025},
  publisher={Nature Publishing Group UK London}
}

@ARTICLE{10255383,
  author={Chen, Genliang and Tang, Shujie and Xu, Shaoqiu and Guan, Tong and Xun, Yuanhao and Zhang, Zhuang and Wang, Hao and Lin, Zhongqin},
  journal={IEEE Transactions on Robotics}, 
  title={Intrinsic Contact Sensing and Object Perception of an Adaptive Fin-Ray Gripper Integrating Compact Deflection Sensors}, 
  year={2023},
  volume={39},
  number={6},
  pages={4482-4499},
  doi={10.1109/TRO.2023.3311610}}

\end{document}